\documentclass[letterpaper, 10 pt, conference]{ieeeconf}

\IEEEoverridecommandlockouts
\usepackage{graphicx}
\usepackage{cuted}
\usepackage{booktabs}
\usepackage[table]{xcolor}
\usepackage{cite}
\usepackage{amsmath}
\usepackage{amssymb}
\usepackage{url}

\newcommand{\prob}{p}
\newcommand{\act}{a}
\newcommand{\obs}{o}
\newcommand{\img}{I}
\newcommand{\tac}{T}
\newcommand{\state}{s}
\newcommand{\lang}{l}

\IEEEaftertitletext{\vspace{-2\baselineskip}}

\title{\LARGE \bf
STAR: Sparse Tactile Representation Learning in Vision–Tactile–Language–Action Models for Dexterous Manipulation}

\author{
Xiangcheng Liu$^{1,*}$, Tianhao Wu$^{2,*}$, Le Zheng$^{2,*}$, Yidong Wang$^{2}$, Bowen Jiang$^{2}$,\\
Mingjie Pan$^{2}$, Xinlin Ren$^{2}$, Yi Liu$^{2}$, and Jianlan Luo$^{1,\dagger}$%
\thanks{$^{1}$Shanghai Innovation Institute; $^{2}$Agibot.}%
\thanks{$^{*}$Equal contribution; $^{\dagger}$Corresponding author.}%
\\[0.3em]
\footnotesize Project page: \url{https://stardex-web.github.io/Star/}%
}

\begin{document}

\maketitle
\thispagestyle{empty}
\pagestyle{empty}

\begin{strip}
    \centering
    \includegraphics[trim=0 2 0 0, clip, width=\textwidth]{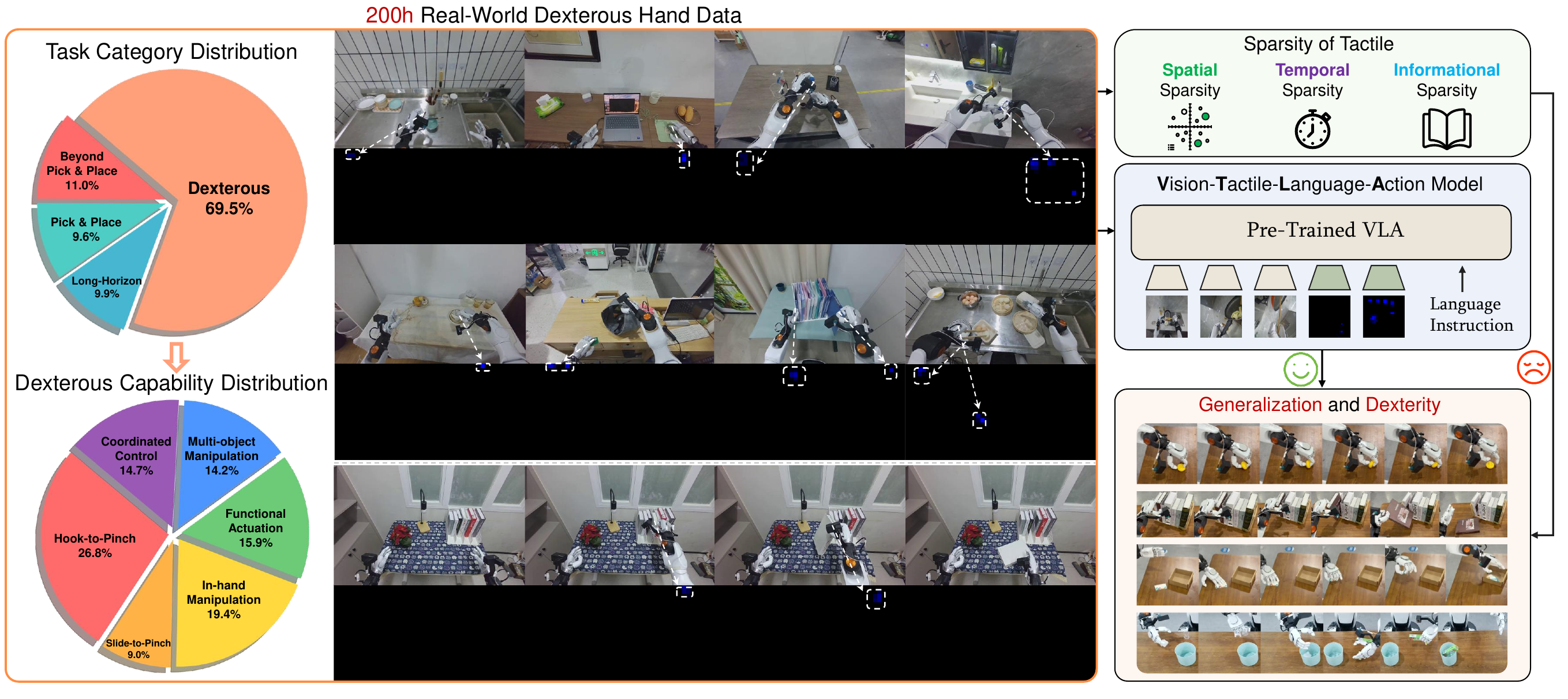}
    \refstepcounter{figure}
    \label{fig:teaser}
    {\footnotesize Fig.~\thefigure. Our training recipe, applied to a 200-hour self-collected dexterous hand dataset with tactile inputs, systematically addresses spatial, temporal, and informational sparsity, enabling both dexterity and generalization across diverse post-training dexterous manipulation tasks.}
\end{strip}

\begin{abstract}

Dexterous manipulation requires coordinated multi-finger control and effective tactile feedback, yet learning these capabilities remains challenging due to the lack of large-scale real-world data and the difficulty of extracting effective representations from sparse tactile signals. We build a robot platform and teleoperation system to collect a 200-hour bimanual dexterous manipulation dataset with synchronized visual, tactile, and language annotations, comprising 10,576 trajectories across 65 tasks, 69.5\% of which involve dexterous multi-finger manipulation. We further propose STAR, an integrated training recipe for vision–tactile–language–action (VTLA) models that addresses the spatial, temporal, and informational sparsity of tactile signals through visual–tactile joint pre-training, sparse-global tactile token representation, and sparse future tactile prediction. Trained on this dataset, STAR achieves a 61\% average success rate across four real-world tasks with 100 post-training trajectories per task, demonstrating dexterous performance under task-specific post-training.

\end{abstract}

\section{INTRODUCTION}

Achieving dexterous manipulation with anthropomorphic robot hands remains a long-standing goal in robotics. While prior works have demonstrated impressive dexterous behaviors on specific tasks~\cite{ye2026visual}, scaling these capabilities to general manipulation settings remains difficult~\cite{liu2025dexndm,Kushal2026simtoolreal,tang2026towards}. Recent dexterous manipulation models have demonstrated generalization across objects and tasks~\cite{bjorck2025gr00t}, but their capabilities remain concentrated on simple skills such as pick-and-place~\cite{beingbeyond2026beingh07,lyu2026lda,zheng2026egoscale}. Moving toward dexterous manipulation requires large-scale dexterous data and learning from sparse tactile signals.

Collecting large-scale real-world dexterous data is challenging because dexterous hands and tactile sensors have limited robustness, and teleoperation is difficult~\cite{xu2025dexumi}. Although egocentric human videos scale easily, the embodiment gap between human and robot hands makes transfer to contact-rich dexterous tasks difficult~\cite{zhang2026unidex}. Moreover, visual observations alone often miss fine-grained contact information. To address this gap, we build a robot and teleoperation system to collect a 200-hour real-world dexterous hand manipulation dataset with synchronized visual, tactile, and language annotations. Importantly, 69.5\% of the trajectories require coordinated multi-finger manipulation that is difficult to achieve with conventional parallel-jaw grippers.
Learning effective tactile representations for vision–tactile–language–action (VTLA) models remains challenging due to the severe sparsity of tactile signals, which appears in several aspects, as shown in Fig.~\ref{fig:teaser}. First, spatial sparsity: for dexterous hands with a large number of tactile sensors, only a small subset of sensors is activated at most timesteps. Second, temporal sparsity: tactile sensing remains inactive during large portions of manipulation, such as when reaching toward an object~\cite{wu2025canonical, li2025adaptive}. Third, informational sparsity: tactile signals provide only local contact information without rich global context. Such severe sparsity increases the difficulty of learning effective tactile representations. Current VLM models rely largely on massive-scale data and relatively weak inductive biases to achieve strong performance. However, scaling dexterous hand tactile data collection to a similar level remains challenging, making stronger inductive biases necessary for effective tactile learning. Existing methods mainly focus on addressing one aspect of tactile sparsity~\cite{zhang2026tacvla,li2026deco,liu2026mla}, but fail to comprehensively address these sparsity challenges. Instead, we propose an integrated training recipe for learning effective tactile representations. First, we apply visual-tactile joint pre-training to prevent the tactile encoder from learning trivial representations caused by spatial sparsity and to facilitate cross-modal alignment. Second, we propose a sparse-global tactile token representation that selects only activated tactile tokens together with learnable global tactile tokens, reducing redundancy while preserving fine-grained information. Finally, we introduce sparse future tactile prediction, leveraging supervision from multiple future horizons to model informative future contact dynamics and enrich tactile representations. By applying this training recipe to our dataset, our policy achieves a 61\% average success rate across 4 real-world dexterous tasks using 100 post-training trajectories per task.

In summary, our contributions are threefold. First, we build a robot and teleoperation system to collect a 200-hour real-world dexterous hand manipulation dataset with tactile sensing, of which 69.5\% of the trajectories require dexterous multi-finger manipulation. Second, we propose an integrated training recipe for dexterous hand VTLA models, combining visual-tactile joint pre-training, a novel sparse-global tactile token representation, and sparse future tactile prediction to learn effective tactile representations. Third, our policy demonstrates dexterous manipulation capabilities across four real-world tasks using 100 post-training trajectories per task.

\section{Related Work}

\subsection{Data Source for Dexterous Hand}
The datasets for dexterous hands can be broadly divided into robot data and non-robot data. Non-robot data includes human data~\cite{punamiya2026egoverse} and wearable device data~\cite{xu2025dexumi}. While such data is easy to scale, it suffers from embodiment gaps in one or more aspects of observations, kinematics, and dynamics, and also lacks commanded actions that contain force information~\cite{wu2025canonical}. As a result, it still requires robot data for alignment and demonstrates limited performance on dexterous and contact-rich tasks~\cite{zheng2026egoscale,beingbeyond2026beingh07}. Robot data includes simulation data~\cite{wu2024dexterous} and real-robot data~\cite{li2026deco}. Although simulation data is easy to scale, it suffers from the sim-to-real gap, since accurately modeling contact-rich interactions, sensor noise, and long-tail real-world scenarios remains extremely challenging. Real-robot teleoperation data does not suffer from these problems, but is difficult to collect due to the limited robustness of dexterous hand hardware and tactile sensors. Moreover, dexterous manipulation data collection requires high-accuracy teleoperation devices, intuitive retargeting, and responsiveness of the whole system. Current real-world dexterous hand datasets~\cite{li2026deco} mainly focus on simple skills such as pick-and-place, with a limited amount of dexterous manipulation data.
\subsection{Dexterous Manipulation}
Task-specific dexterous manipulation has been widely studied, with many skills demonstrated~\cite{wu2025canonical}. However, these methods typically require task-specific design or a separate model per task. To improve generalization, prior works imitate human reference trajectories~\cite{Kushal2026simtoolreal}, but they require accurate references and poorly capture dynamics. Recent works explore foundation models, yet scarce real-world dexterous data forces them to rely on human data or mixtures of human and robot data~\cite{zheng2026egoscale,beingbeyond2026beingh07, lyu2026lda} for pre-training, followed by post-training on real-robot data. However, embodiment gaps, missing tactile modalities, and absent commanded actions enlarge the human-robot distribution gap as tasks become more dexterous and contact-rich. Consequently, current methods achieve high success mainly on simple tasks such as pick-and-place, while showing limited performance on dexterous tasks~\cite{zheng2026egoscale,zhang2026unidex}. In contrast, our VTLA model is finetuned on real-robot data rich in dexterous manipulation, supporting adaptation to diverse real-world tasks with 100 task-specific trajectories.

\subsection{Vision-Tactile-Language-Action Models}
Temporal and informational sparsity exist in both parallel grippers and dexterous hands. For temporal sparsity, current methods introduce gating mechanisms, either heuristic~\cite{zhang2026tacvla} or adaptive~\cite{li2026vla}, to inject all tactile tokens at appropriate times. For informational sparsity, world-model-based methods predict continuous future tactile frames to model informative future contact dynamics, thereby enriching tactile representations to some extent. Spatial sparsity is more severe in dexterous hands. Compared with parallel grippers, which only contain two tactile sensor pads, dexterous hands can be covered with a large area of tactile sensors, leading to severe token redundancy. Current methods~\cite{li2026deco,tang2026towards} divide tactile signals into several regions, which reduce redundancy at the raw data level. To facilitate the use of tactile, cross-modal alignment~\cite{liu2026mla} is also necessary before policy training. While current works mainly focus on addressing a single aspect, we propose an integrated training recipe from tactile encoder pre-training to policy training that comprehensively addresses tactile sparsity.

\section{Problem Statement}

We focus on the problem of modeling the distribution of $\prob(\act_t|\obs_t)$, where each observation $\obs_t = \{\img_t, \tac_t, \state_t, \lang_t \}$. Here, $\img_t$ consists of multiple images $\{ \img^1_t, ..., \img^N_t \}$, and $\tac_t$ consists of multiple tactile signals $\{ \tac^1_t, ..., \tac^M_t \}$. $\state_t$ represents the robot proprioceptive state, while $\lang_t$ represents the language instruction. $\act_t$ denotes an action chunk of future actions, defined as $\act_t = \{ \act^1_t, \act^2_t, ..., \act^H_t \}$. Both the state $\state_t$ and the action $\act_t$ consist of the robot arm end-effector pose and hand joint positions. We represent both $\state_t$ and $\act_t$ relative to the current robot state $\state_t$. 

We formulate action generation as a conditional flow matching process. 
We train a velocity field $\mathbf{v}_\theta(\act_t^\tau, \obs_t, \tau)$ to transport samples from a Gaussian prior $\boldsymbol{\epsilon} \sim \mathcal{N}(\mathbf{0}, \mathbf{I})$ to the target action distribution $\act_t \sim \prob(\act_t|\obs_t)$:

\begin{equation}
\mathcal{L}_{\mathrm{FM}}(\theta)
=
\mathbb{E}_{\act_t,\boldsymbol{\epsilon},\tau}
\left[
\left\|
\mathbf{v}_\theta(\act_t^\tau, \obs_t, \tau)
-
(\act_t - \boldsymbol{\epsilon})
\right\|_2^2
\right].
\end{equation}
The interpolated action is defined as:
\begin{equation}
\label{eq:flow-interpolation}
\act_t^\tau
= (1-\tau)\boldsymbol{\epsilon} + \tau \act_t.
\end{equation}

We sample a timestep $\tau \in [0,1]$ from a shifted beta distribution. During inference, we encode the current observation $\obs_t$ as prefix tokens and cache their key-value states. The model then integrates the learned velocity field from an initial noisy action sample using 10 forward Euler steps:
\begin{equation}
\label{eq:es}
\act_t^{\tau+\delta}
=
\act_t^\tau
+
\delta\,
\mathbf{v}_\theta(\act_t^\tau, \obs_t, \tau),
\end{equation}
where $\delta$ is the integration step size.

\section{Method}
Due to the lack of a high-quality dexterous hand dataset that includes dexterous finger coordination and tactile modality, we first carefully build a robot and teleoperation system to collect a 200-hour real-world dataset (Sec.~\ref{sec:dataset}). To effectively use the tactile modality in such a dataset, we propose an integrated training recipe (Sec.~\ref{sec:recipe}) to comprehensively address the sparsity of tactile signals.
\subsection{Real-World Dexterous Manipulation Dataset}\label{sec:dataset}
\begin{figure}[t]
    \centering
    \includegraphics[trim=0 0 0 0, clip, width=0.95\linewidth]{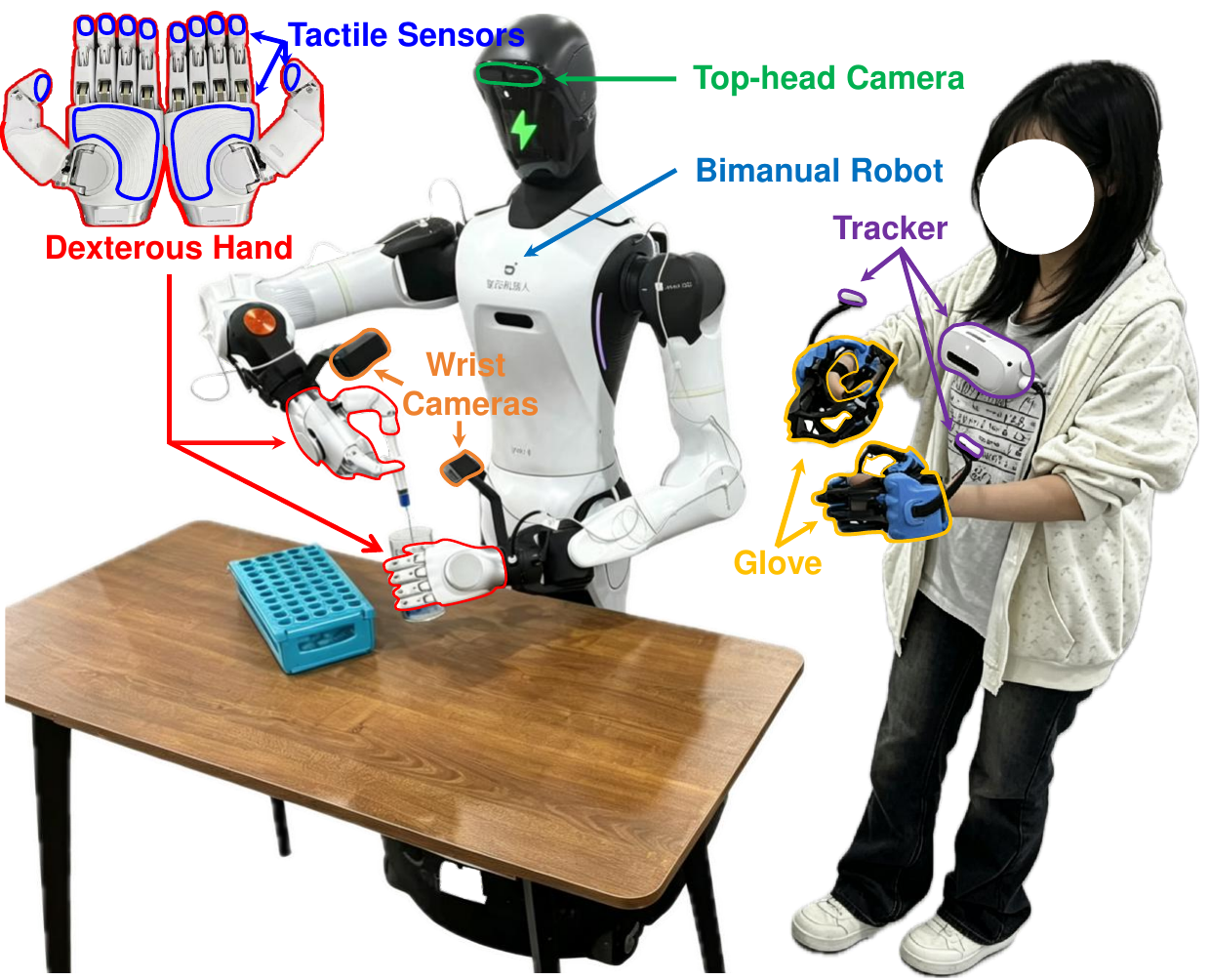}
    \caption{Real Robot and Teleoperation System.}
    \label{fig:system}
\end{figure}
To support large-scale data collection, dexterous hands should have good proprioceptive consistency, avoid overheating during long-term operation, and provide strong impact resistance, while offering additional degrees of freedom for dexterous manipulation and durable large-area tactile sensing. Based on these considerations, we choose 10-DoF dexterous hands integrated with 268-dimensional piezoresistive tactile sensors, mounted on a bimanual wheeled mobile robot, as shown in Fig.~\ref{fig:system}. Another critical challenge in collecting dexterous manipulation datasets is teleoperation. The accuracy of the teleoperation device, the effectiveness of the retargeting algorithm, and the responsiveness of the system largely affect both manipulation difficulty and data quality. We choose a skeleton-based glove as the teleoperation device to capture nuanced human motions. To retarget human commands to the dexterous hands, we use the DexPilot algorithm~\cite{handa2020dexpilot}, while adjusting the fingertip positions of the glove URDF to match the dexterous hands and adding orientation constraints for more intuitive control. More importantly, we disable the automatic finger close mechanism, since it introduces difficulties for fine-grained finger motion control. To further improve teleoperation responsiveness, we implement the algorithm in C++ and integrate it into a multiprocessing pipeline.
With the carefully optimized robot and teleoperation system, we collect a 200-hour real-world dexterous manipulation dataset. Notably, 69.5\% of the collected trajectories require dexterous manipulation. Another 20\% consist of other primitive skills such as pick-and-place and insertion, while the remaining 10\% are long-horizon manipulation trajectories composed of primitive skills. The dataset has 10k+ trajectories, with a total of 65 tasks, collected in real-world scenarios, as shown in Fig.~\ref{fig:teaser}. The dataset includes images from left and right wrist cameras $\{\img^1_t, \img^2_t \}$ and a top-head camera $\{ \img^3_t \}$, tactile signals from left and right hands $\{ \tac^1_t, \tac^2_t \}$, robot states $\state_t$, command actions $\act_t$, and manually annotated task-level language instructions $\lang_t$. All data are synchronized to 30 Hz according to timestamps. Additional dataset, tactile representation, and evaluation details are provided in the appendix.
\subsection{Training Recipe}\label{sec:recipe}
\begin{figure*}[t]
    \centering
    \includegraphics[trim=0 0 0 0, clip, width=1.0\textwidth]{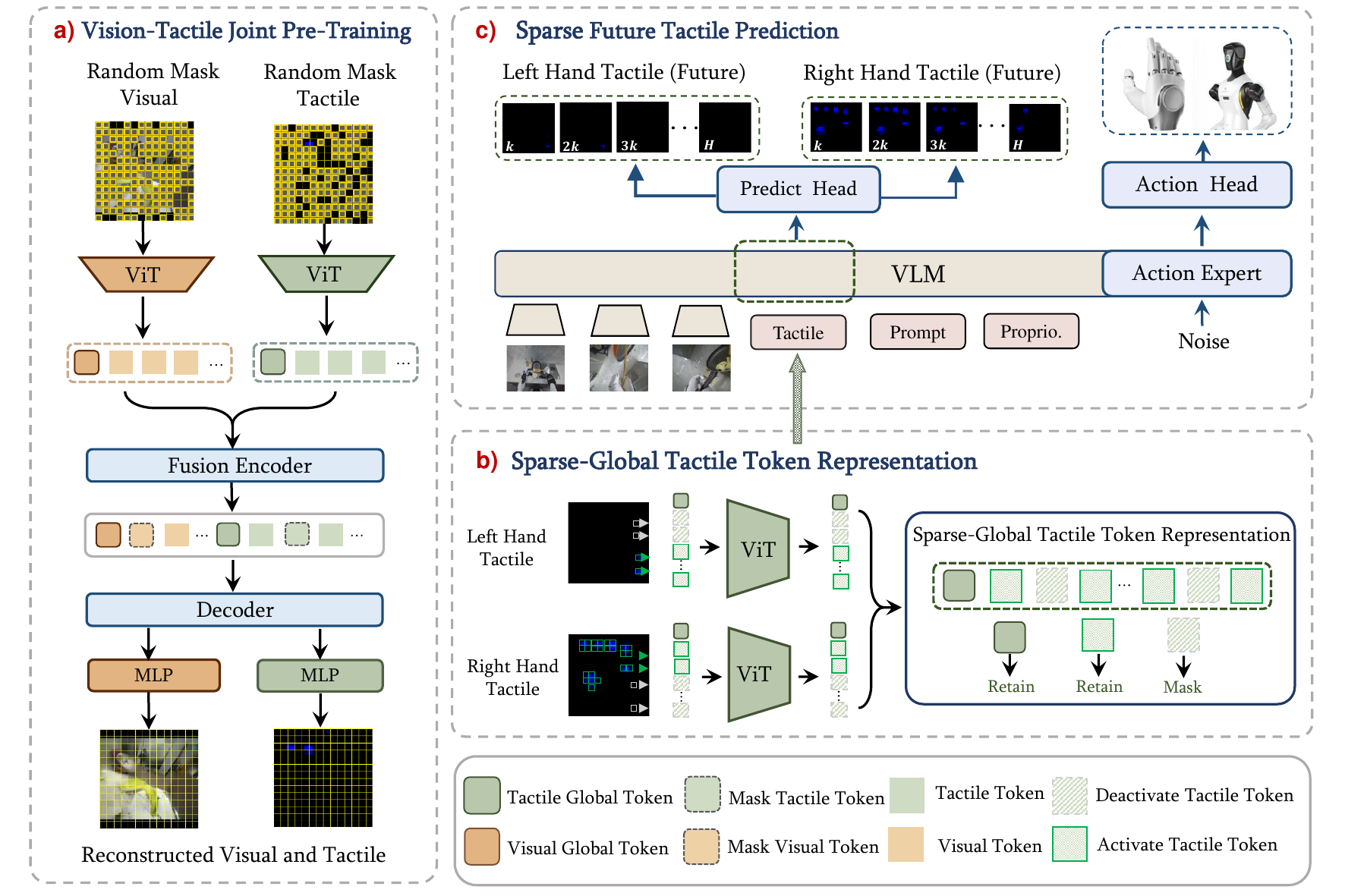}
    \caption{\textbf{Training Recipe}. a) We jointly predict the masked tokens of RGB and tactile images for both intra- and inter-modal tactile pre-training. b) We construct the sparse-global tactile token representation for policy training by retaining activated spatial tactile tokens and per-hand global tactile tokens. c) During policy pre-training, we add sparse future tactile prediction to model informative future contact dynamics.}
    \label{fig:pipeline}
\end{figure*}
A dexterous hand covered with a large area of tactile sensors can lead to highly sparse inputs. By analyzing our collected dataset, we observe that the highest activated taxel ratio across all frames is only 24\%. Meanwhile, the activated tactile frame number ratio is only 35\%, indicating severe temporal sparsity. On the other hand, the tactile modality itself only represents local contact information without global context information, which further leads to informational sparsity.
To address the problem, we first pre-train the tactile encoder with visual-tactile joint pre-training (Sec.~\ref{sec:jointpre}). For policy training, we initialize the model with the $\pi_{0.5}$ weights, except for the action layer to match the action dimension of our robot. We finetune the model using our dataset, and then use task-specific data for post-training.

\subsubsection{Visual-Tactile Joint Pre-training} \label{sec:jointpre}
\begin{figure}[t]    \centering    \includegraphics[trim=0 0 0 0, clip, width=0.8\linewidth]{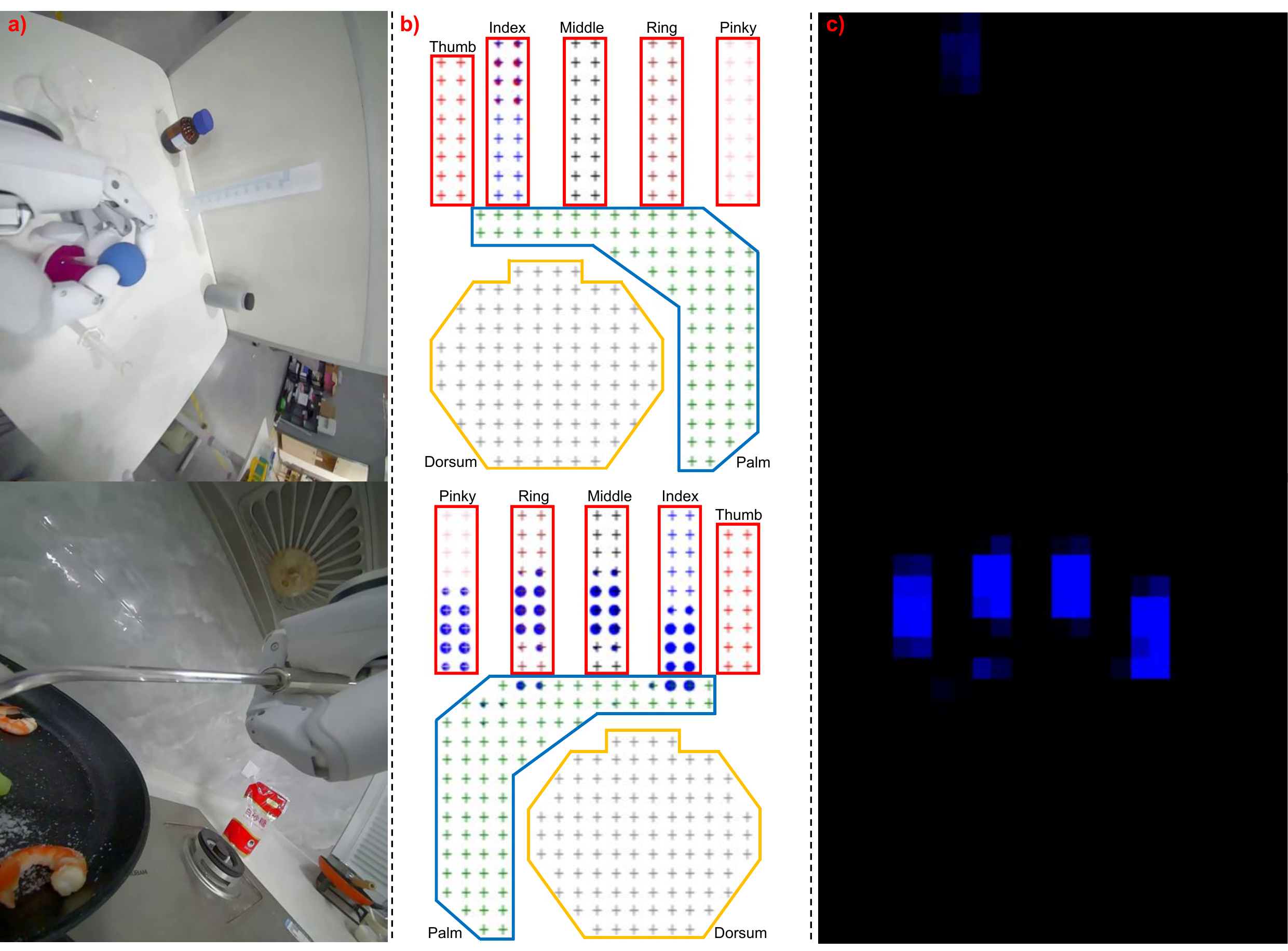}    \caption{Visualization of Tactile Representation: a) RGB image. b) Spatial Relationship of Taxels. c) Tactile Representations.    }    \label{fig:tacrep}\end{figure}

Due to the lack of tactile encoders pre-trained on large-scale dexterous manipulation datasets, we pre-train the tactile encoder using our self-collected dataset. Following T-Dex~\cite{guzey2023dexterity}, we represent tactile signals as images. Fig.~\ref{fig:tacrep} visualizes representative RGB observations, the spatial taxel layout, and the resulting tactile representations. The 268 taxels are distributed on the palm and inner finger contact surfaces, with the finger edge regions left uncovered; non-taxel regions in the tactile images are filled with zeros. Due to the spatial sparsity of tactile signals, pre-training solely on the tactile modality can cause the model to overfit non-activated tactile values, resulting in trivial tactile representations. Therefore, we apply visual-tactile joint pre-training following M2VTP~\cite{liu2024masked}, as shown in Fig.~\ref{fig:pipeline}(a), where image tokens and tactile tokens are concatenated together for joint masked token prediction. We construct the pair of corresponding wrist image and tactile signals $(\img_i, \tac_i), i=1,2,$ for training, using the same encoder for left and right hand pairs. Under joint visual-tactile pre-training, trivially predicting all-zero tactile signals becomes suboptimal, as it is inconsistent with the visual context. This encourages the tactile encoder to extract more informative tactile representations, facilitating both understanding of the inherent structure of tactile signals and alignment with vision. Both RGB images and tactile images mask 75\% of the patches, and we use patch-wise normalized loss for supervision:
\begin{equation}
\mathcal{L}_{\mathrm{joint}}
= \sum_{m \in \{\mathrm{img}, \mathrm{tac}\}} \frac{1}{|P^{\mathrm{mask}}_{m}|}
\sum_{j \in P^{\mathrm{mask}}_{m}}
\left\| \hat{\mathbf{x}}_m^{(j)} - \bar{\mathbf{x}}_m^{(j)} \right\|^2_2.
\end{equation}
The normalized target patch is defined as:
\begin{equation}
\label{eq:patch-normalization}
\bar{\mathbf{x}}_m^{(j)}
=
\frac{\mathbf{x}_m^{(j)} - \mu_m^{(j)}}{\sigma_m^{(j)}}.
\end{equation}
Here, $\hat{\mathbf{x}}_m^{(j)}$ and $\bar{\mathbf{x}}_m^{(j)}$ denote the predicted and normalized ground-truth patches for the $j$-th masked patch of modality $m$, respectively. $\mathbf{x}_m^{(j)}$ is the original patch, while $\mu_m^{(j)}$ and $\sigma_m^{(j)}$ represent the mean and standard deviation of the pixels within the $j$-th patch of modality $m$. $P^{\mathrm{mask}}_{m}$ represents the set of masked patch indices for modality $m$, with $|P^{\mathrm{mask}}_{m}|$ denoting its cardinality.

\subsubsection{Sparse-Global Tactile Token Representation}
\label{sec:rep}
Due to the severe sparsity of tactile, directly using all tactile tokens introduces redundancy from inactive tactile regions, similar to the token redundancy observed in vision transformers~\cite{rao2021dynamicvit}.
We propose a sparse-global tactile token representation to address the problem, as shown in Fig.~\ref{fig:pipeline}(b). The tactile encoder outputs spatial patch tokens together with a per-hand learnable global tactile token that summarizes the full tactile field. We construct a contact gate from calibrated tactile readings and treat a spatial token as activated when any taxel in the token exceeds a sensor-specific contact threshold. This gate depends only on contact activation, and can be instantiated with contact or change maps from other tactile sensors such as GelSight. We apply the gate only to spatial patch tokens, retaining activated tokens and masking inactive ones, while always keeping the global tactile token.
In the VLM backbone, this global tactile token is prevented from attending to visual, language, and action tokens, while these tokens are allowed to attend to it. This asymmetric attention keeps the global token tactile-specific while allowing other modalities to access tactile information. This hybrid representation filters inactive tactile tokens while preserving fine-grained tactile details and global contact information.

\subsubsection{Sparse Future Tactile Prediction}\label{sec:futurepredict}Historical tactile observations provide only limited information about the current contact state, whereas anticipating how contact may evolve in the future is critical for dexterous manipulation~\cite{li2025adaptive}. Therefore, we introduce sparse future tactile prediction as an auxiliary task during both policy finetuning and post-training. As shown in Fig.~\ref{fig:pipeline}(c), under a fixed budget of five predicted tactile frames, we uniformly distribute the prediction targets across the action horizon at timesteps $T_{\mathrm{S}} = \{10, 20, 30, 40, 50\}$, where $H{=}50$. This enables the model to capture longer-horizon contact dynamics through temporally distributed future supervision. We add MLPs to predict future force for left and right hands, respectively, from the output tactile tokens of VLM. The patch-wise prediction loss is defined as:
\begin{equation}
\mathcal{L}_{future} = \frac{1}{|T|} \sum_{t \in T} \frac{1}{|P^{t}_{tac}|} \sum_{j \in P^{t}_{tac}} \left\| \hat{\mathbf{x}}^{(j)}_{tac} - \mathbf{x}^{(j)}_{tac} \right\|_2^2,
\end{equation}
where $\hat{\mathbf{x}}^{(j)}_{tac}$ and $\mathbf{x}^{(j)}_{tac}$ represent the predicted and ground-truth patches for the $j$-th tactile patch, respectively. $P^{t}_{tac}$ represents the set of tactile patch indices at future timestep $t$, with $|P^{t}_{tac}|$ denoting its cardinality. By predicting tactile observations across multiple future horizons, the model is encouraged to learn representations that encode informative future contact dynamics. This auxiliary objective enriches tactile representations beyond what can be inferred from historical observations alone, thereby alleviating the informational sparsity.

\section{Experiments}\label{sec:exp}
We conduct comprehensive real-world experiments to answer the following questions:
\begin{itemize}    \item Can our real-world dexterous manipulation data improve performance?     \item Does our proposed training recipe address the sparsity of the tactile modality?    \item How does the tactile modality affect performance?
\end{itemize}

\subsection{Experiment Setup}
\begin{figure*}[t]
    \centering
    \includegraphics[trim=0 0 0 0, clip, width=\textwidth]{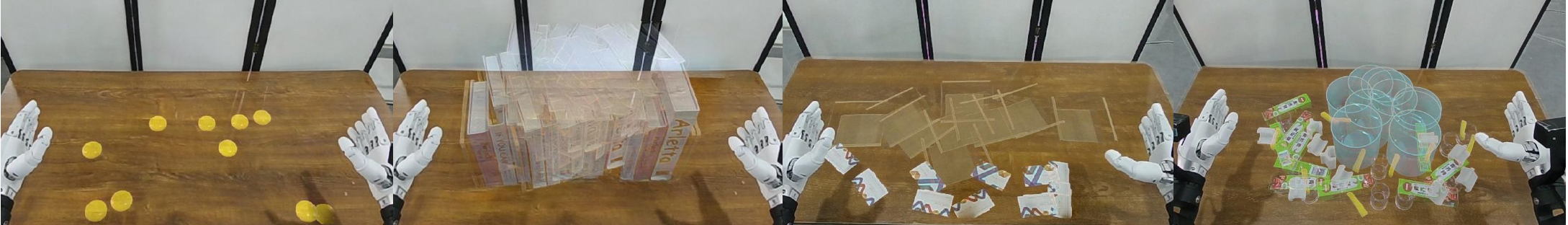}
    \caption{Visualization of Evaluation Setup.}
    \label{fig:evalsetup}
\end{figure*}

We evaluate our policy on four dexterous manipulation tasks, as shown in Fig.~\ref{fig:evalsetup}: \textbf{Earbud flipping:} This task requires the robot to use the thumb and index finger to flip the earbud by 180 degrees. \textbf{Stacked-book retrieval:} Due to tight stacking, this task requires the robot to hook the upper part of the book with the index finger, and then grasp it using the thumb and middle finger to retrieve the target book. \textbf{Postcard retrieval:} This task requires the robot to slide the postcard to the edge of the table, then pick and place it into the box. \textbf{Multi-object grasping:} This task requires each hand to sequentially collect two pieces of trash using different finger groups and discard them into the trash can.

We use two metrics to evaluate policy performance: \textbf{success rate (SR):} the ratio between the number of successful trials and the total number of trials. \textbf{task completion rate (TCR):} the ratio between the number of completed subtasks and the total number of subtasks. To compute TCR, each task is divided into subtasks: grasp/rotate/release for earbud flipping, pull/grasp/retrieve/place for stacked-book retrieval, slide/grasp/place for postcard retrieval, and per-object grasp and disposal for multi-object grasping, resulting in eight subtasks for this task.
We compare our policy with the following baselines: \textbf{GR00T N1.7:} this baseline~\cite{bjorck2025gr00t} is pre-trained on a mixture of diverse real robot data, synthetic data, egocentric data, and internet videos. \textbf{LDA-1B:} this baseline~\cite{lyu2026lda} is pre-trained on 30k hours of human and cross-embodiment robot data. \textbf{$\boldsymbol{\pi_{0.5}}$:} this baseline~\cite{black2025pi_} is pre-trained on large-scale cross-embodiment parallel gripper data without dexterous hand datasets. All baselines are initialized from the officially released pretrained weights for post-training. We use the officially recommended finetuning hyperparameters for each baseline when available, while keeping the same 100 task-specific trajectories per task, post-training steps, and evaluation protocol across methods. For \textit{LDA-1B} and \textit{$\pi_{0.5}$}, we reinitialize the action layer to match our action dimension. The objects used in the four evaluation tasks are held out from the 200-hour pre-training dataset. During task-specific post-training, each policy is adapted with 100 trajectories using the evaluation-task objects, and is then tested on held-out initial configurations. For all evaluations, each method is evaluated over 20 trials per task. We use the same randomly selected initial scene configurations across all methods to ensure a fair comparison.

\subsection{Comparison of Different Manipulation Policies}
\begin{table*}[t]
    \centering
    \caption{Comparison of Baselines.}
    \label{table:main}
    \scriptsize
    \resizebox{0.9\textwidth}{!}{%
        \begin{tabular}{c|cc|cc|cc|cc|cc}
\toprule
\textbf{Method}                 
& \multicolumn{2}{c|}{\textbf{Earbud flipping}} 
& \multicolumn{2}{c|}{\textbf{Stacked-book retrieval}} 
& \multicolumn{2}{c|}{\textbf{Postcard retrieval}}     
& \multicolumn{2}{c|}{\textbf{Multi-object grasping}}    
& \multicolumn{2}{c}{\textbf{Avg}} \\ 

\cmidrule(lr){2-3}
\cmidrule(lr){4-5}
\cmidrule(lr){6-7}
\cmidrule(lr){8-9}
\cmidrule(lr){10-11}

& \textbf{SR} & \textbf{TCR}
& \textbf{SR} & \textbf{TCR}
& \textbf{SR} & \textbf{TCR}
& \textbf{SR} & \textbf{TCR}
& \textbf{SR} & \textbf{TCR} \\

\midrule

$\textbf{GR00T N1.7}$        
& 0.00 & 0.00
& 0.00 & 0.00
& 0.00 & 0.00
& 0.00 & 0.00
& 0.00 & 0.00 \\

\textbf{LDA-1B}                 
& 0.05 & 0.05
& 0.00 & 0.09
& 0.15 & 0.22
& 0.00 & 0.02
& 0.05 & 0.09 \\

$\textbf{GR00T N1.7-Dex}$        
& 0.00 & 0.00
& 0.05 & 0.05
& 0.25 & 0.40
& 0.00 & 0.14
& 0.08 & 0.15 \\

\textbf{LDA-1B-Dex}                 
& 0.40 & 0.40
& 0.00 & 0.03
& 0.30 & 0.40
& 0.00 & 0.11
& 0.18 & 0.23 \\

$\boldsymbol{\pi_{0.5}}$        
& 0.15 & 0.17
& 0.35 & 0.49
& 0.60 & 0.73
& 0.00 & 0.36
& 0.28 & 0.44 \\

\textbf{$\pi_{0.5}$-Dex w/o STAR}      
& 0.25 & 0.40
& 0.45 & 0.52
& 0.75 & 0.83
& 0.30 & 0.69
& 0.44 & 0.61 \\

\rowcolor{gray!20}\textbf{Ours} 
& \textbf{0.65} & \textbf{0.73}
& \textbf{0.55} & \textbf{0.83}
& \textbf{0.85} & \textbf{0.90}
& \textbf{0.40} & \textbf{0.72}
& \textbf{0.61} & \textbf{0.79} \\ 

\bottomrule
\end{tabular}
    }
\end{table*}

As shown in Tab.~\ref{table:main}, \textit{Ours} achieves the highest SR and TCR across all tasks, indicating the effectiveness of our method. For \textit{GR00T N1.7}, we report zero success because, starting from the official pretrained weights and using the same post-training steps as other baselines, the policy showed poor convergence and could not complete any trial safely. During evaluation, severe arm oscillations caused unsafe motions that could lead to collisions with the robot or the environment, and these trials were terminated for safety. Although \textit{LDA-1B} is pre-trained on large-scale human and robot data, its performance is relatively low, even worse than \textit{$\pi_{0.5}$}. We observe that many failures of \textit{LDA-1B} arise from inaccurate interaction poses. This may be because training on mixed human and robot data, together with the latent training paradigm, biases the model toward high-level representations while sacrificing the fine-grained spatial precision required for dexterous manipulation.
We further finetune the \textit{GR00T N1.7} and \textit{LDA-1B} with our self-collected dataset, denoted as \textit{GR00T N1.7-Dex} and \textit{LDA-1B-Dex}. We also include \textit{$\pi_{0.5}$-Dex w/o STAR}, which denotes $\pi_{0.5}$ finetuned on our self-collected dexterous dataset without the proposed STAR training recipe. The performance of both \textit{GR00T N1.7} and \textit{LDA-1B} improves after finetuning, validating the effectiveness of our dataset. However, \textit{GR00T N1.7-Dex} and \textit{LDA-1B-Dex} still fall short of \textit{$\pi_{0.5}$}, indicating that large-scale pretraining alone does not fully close the embodiment and modality gaps for contact-rich dexterous manipulation.
\textit{$\pi_{0.5}$} achieves both high SR and TCR. Although \textit{$\pi_{0.5}$} has not been pretrained on dexterous hand data, it is pre-trained on diverse real-robot data and therefore learns a strong prior for arm motion, which is critical for reaching precise locations. However, there is a large embodiment gap between dexterous hands and grippers, which limits its performance. It also lacks tactile input.

\subsection{Importance of Training Recipe}
\begin{table*}[t]
    \centering
    \caption{\textbf{Ablation of Our Training Recipe.}
    VTP: visual-tactile joint pre-training.
    STTR: sparse tactile token representation.
    GTTR: global tactile token representation.
    SFTP: sparse future tactile prediction.
    DFTP: dense future tactile prediction.
    \textbf{Avg.}: average over the two tasks.}
    \label{table:ablation}
    \scriptsize
    \setlength{\tabcolsep}{8pt}
    \begin{tabular}{c|cc|cc|cc}
\toprule
\textbf{Method}                 
& \multicolumn{2}{c|}{\textbf{Earbud flipping}} 
& \multicolumn{2}{c|}{\textbf{Stacked-book retrieval}}    
& \multicolumn{2}{c}{\textbf{Avg.}} \\
\cmidrule(lr){2-3}\cmidrule(lr){4-5}\cmidrule(lr){6-7}
& \textbf{SR} & \textbf{TCR}
& \textbf{SR} & \textbf{TCR}
& \textbf{SR} & \textbf{TCR} \\
\midrule
\textbf{Ours w/o STAR}           
& 0.25 & 0.40
& 0.45 & 0.52
& 0.35 & 0.46 \\

\textbf{Ours w/o VTP}           
& 0.30 & 0.42
& 0.45 & 0.48
& 0.38 & 0.45 \\

\textbf{Ours w/o STTR}         
& 0.40 & 0.52
& 0.25 & 0.36
& 0.33 & 0.44 \\

\textbf{Ours w/o GTTR}           
& 0.25 & 0.33
& 0.35 & 0.48
& 0.30 & 0.41 \\

\textbf{Ours w/o SFTP}          
& 0.55 & 0.62
& 0.50 & 0.76
& 0.53 & 0.69 \\

\textbf{Ours w/o SFTP, w/ DFTP}          
& 0.55 & 0.62
& 0.40 & 0.63
& 0.48 & 0.62 \\

\rowcolor{gray!20}
\textbf{Ours} 
& \textbf{0.65} & \textbf{0.73}
& \textbf{0.55} & \textbf{0.83}
& \textbf{0.60} & \textbf{0.78} \\ 

\bottomrule
\end{tabular}

\end{table*}
    To validate the importance of the training recipe, we conduct ablation studies on two representative tasks with complementary tactile requirements: \textit{Earbud flipping}, where fingertip contact is frequent and tactile feedback is critical for successful reorientation, and \textit{Stacked-book retrieval}, where the task relies more on visual localization and geometric motion planning, with relatively sparse informative tactile feedback. This choice allows us to evaluate whether each component of STAR contributes under both tactile-rich and tactile-sparse manipulation regimes, while keeping the real-robot ablation cost manageable. All ablation variants are independently trained using the same two-stage protocol as Ours: policy pre-training on the full 200-hour dataset followed by task-specific post-training on the same 100 trajectories per task. For each ablation, the corresponding component is removed or replaced before the 200-hour policy pre-training stage, and the resulting configuration is retained during the subsequent task-specific post-training stage; all other training settings are kept identical. Specifically, we conduct the following ablation studies: \textit{Ours w/o STAR} directly incorporates tactile inputs into the policy without visual–tactile joint pre-training, sparse-global tactile token representation, or sparse future tactile prediction. All tactile tokens are directly fed into the policy, and the tactile encoder is trained end-to-end during policy training. \textit{Ours w/o VTP} does not pre-train the tactile encoder and directly uses tactile inputs for policy training. Within the sparse-global tactile token representation (SGTTR), we ablate its sparse tactile token representation (STTR) and global tactile token representation (GTTR) components. \textit{Ours w/o STTR} uses all tactile tokens, including both activated and deactivated tactile tokens, together with the global tactile tokens as input to the VLM. \textit{Ours w/o GTTR} masks the global tactile tokens, only uses activated tactile tokens as input to the VLM. \textit{Ours w/o SFTP} does not predict future tactile signals during policy training. \textit{Ours w/o SFTP, w/ DFTP} replaces SFTP with dense future tactile prediction. Both variants predict exactly five frames: DFTP uses $T_{\mathrm{D}}=\{1,2,3,4,5\}$, whereas SFTP uses $T_{\mathrm{S}}=\{10,20,30,40,50\}$.
As shown in Tab.~\ref{table:ablation}, compared with \textit{Ours}, eliminating any component of the training recipe degrades the performance of the policy, indicating that integrating all three components better addresses the sparsity issue. VTP helps align visual and tactile information, enabling the policy to better utilize both modalities. Without such pre-training, the performance also drops significantly. Without STTR, the policy needs to learn to extract useful features from tokens containing a large amount of redundant information, which limits its ability to effectively utilize all the modalities. Without GTTR the performance also decreases significantly, demonstrating the necessity of introducing global tactile tokens to aggregate global tactile information. Comparing \textit{Ours w/o STTR} with \textit{Ours w/o GTTR}, although their overall SR and TCR are similar, their performance differs across tasks. For the stacked-book retrieval task, which relies less on tactile information (Sec.~\ref{sec:tac-role}), STTR is more beneficial by reducing tactile token redundancy. In contrast, for the earbud flipping task, which relies more on tactile information (Sec.~\ref{sec:tac-role}), GTTR plays a more important role by aggregating global tactile information. Together with visual-tactile alignment, these complementary designs help the policy utilize tactile information. Therefore, eliminating SFTP does not significantly affect performance, while the improvement in SR and TCR indicates that enriching tactile representations remains effective.

Finally, we compare two temporal allocations under the same prediction budget of five tactile frames. At the 30 Hz control frequency, DFTP concentrates its targets within the next 0.17 s, whereas SFTP distributes the same number of targets over 1.67 s. Replacing SFTP with DFTP decreases performance on both tasks, showing that, under a fixed target budget, distributing supervision across the action horizon is more effective than concentrating it in the immediate future. We hypothesize that adjacent near-term tactile prediction is easier to fit and therefore provides less information about contact transitions. Since sampling interval and maximum temporal extent are coupled in this comparison, the ablation does not independently attribute the gain to either factor.

\subsection{Role of Tactile}
\label{sec:tac-role}
\begin{table}[t]
    \centering
    \caption{\textbf{Ablation of Tactile.}
    T: tactile.
    \textbf{Avg.}: average over the two tasks.}
    \label{table:ablation_role_of_tactile}
    \scriptsize
    \begin{tabular}{c|cc|cc|cc}
\toprule
\textbf{Method}                 
& \multicolumn{2}{c|}{\textbf{Earbud flipping}} 
& \multicolumn{2}{c|}{\textbf{Stacked-book retrieval}}    
& \multicolumn{2}{c}{\textbf{Avg.}} \\

\cmidrule(lr){2-3}
\cmidrule(lr){4-5}
\cmidrule(lr){6-7}

& \textbf{SR} & \textbf{TCR}
& \textbf{SR} & \textbf{TCR}
& \textbf{SR} & \textbf{TCR} \\

\midrule

\textbf{Ours w/o T}             
& 0.35 & 0.52
& 0.55 & 0.71
& 0.45 & 0.61 \\

\textbf{Ours mask T}          
& 0.55 & 0.67
& 0.50 & 0.71
& 0.53 & 0.69 \\

\rowcolor{gray!20}
\textbf{Ours} 
& \textbf{0.65} & \textbf{0.73}
& \textbf{0.55} & \textbf{0.83}
& \textbf{0.60} & \textbf{0.78} \\ 

\bottomrule
\end{tabular}

\end{table}

\begin{figure*}[t]
    \centering
    \includegraphics[trim=0 0 0 0, clip, width=0.8\textwidth]{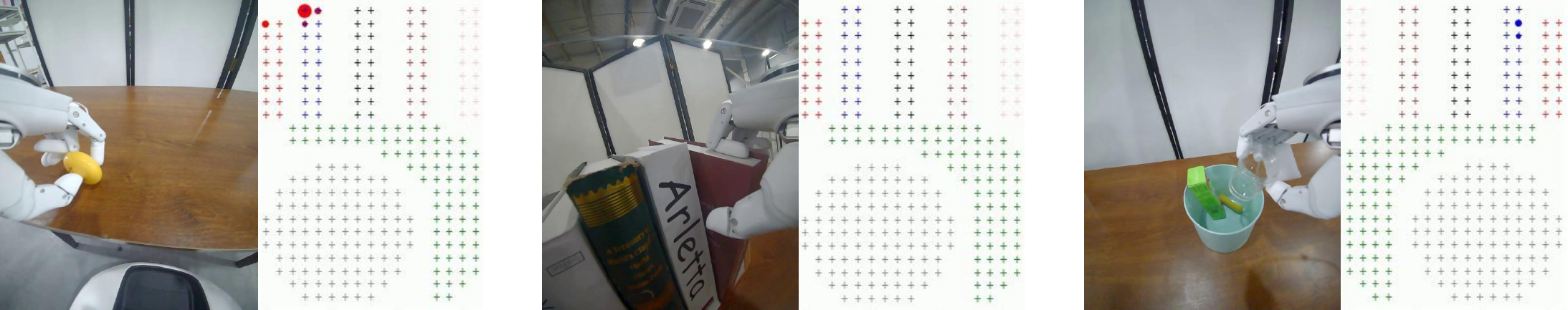}
    \caption{\textbf{Visualization of Tactile During Policy Rollout.}}
    \label{fig:tacvis}
\end{figure*}

To analyze the role of tactile, we conduct the following ablation studies: \textit{Ours w/o T} finetunes $\pi_{0.5}$ with our 200-hour dataset but without tactile as input. \textit{Ours mask T} masks the tactile input during policy inference, following~\cite{wu2025canonical}. As shown in Tab.~\ref{table:ablation_role_of_tactile}, the overall SR and TCR of \textit{Ours w/o T} are both higher than those of \textit{$\pi_{0.5}$}, demonstrating the effectiveness of our dataset. Comparing \textit{Ours} with \textit{Ours w/o T}, while the overall SR and TCR of \textit{Ours} are higher, the improvement mainly stems from the earbud flipping task. By analyzing the policy rollouts, we attribute this phenomenon to the following reasons. Since our policy encourages the utilization of tactile information, it may focus more on tactile signals at the expense of effectively utilizing visual information. However, the current tactile sensors on our dexterous hand do not cover the finger edges, which limits the policy's ability to use edge contact information, as shown in Fig.~\ref{fig:tacvis}. This also illustrates the reason that for the multi-object grasping task, the TCR of \textit{Ours} is much higher than the SR. This gap is mainly due to the slight deviations when grasping with the finger edges of the little and ring fingers. In contrast, for the earbud flipping task, the policy uses the fingertips, which are covered with tactile sensors, thereby improving performance. Tactile masking results further show that removing tactile inputs degrades performance on both tasks, indicating that our policy uses tactile for action generation.

\section{Limitations}
Currently, our dataset only covers a limited set of tasks, and its size is still small compared to parallel gripper datasets, which can be further extended to more diverse tasks and larger-scale data collection for training a truly dexterous hand foundation model. Our training recipe currently only includes visual-tactile alignment before policy training, while alignment with language is also important. We also observe that, for some tasks, using tactile does not bring positive effects. This may be due to the policy focusing more on tactile signals at the expense of effectively utilizing visual information, along with the limitation of tactile sensor coverage. In addition, extending the method to other hands and tactile sensors remains an important direction, and future work should further improve its robustness under sensor noise, drift, and missing taxels.

\section{CONCLUSIONS}

In this work, we build a robot and teleoperation system to collect a 200-hour real-world dexterous hand dataset with tactile modality, where 69.5\% of the trajectories involve dexterous manipulation, helping mitigate the scarcity of high-quality, multi-modal real-world dexterous manipulation data. By analyzing the dataset, we find that the challenge of using tactile as additional input for VLA models mainly arises from the spatial, temporal, and informational sparsity of tactile signals. Therefore, we propose an integrated training recipe that combines visual-tactile joint pre-training, sparse-global tactile token representation, and sparse future tactile prediction to systematically address tactile sparsity. By applying our method to the self-collected dataset, our policy demonstrates dexterous performance across four real-world tasks with 100 post-training trajectories per task.

\bibliographystyle{IEEEtran}
\bibliography{ref}

\appendix
\subsection{Real Robot and Teleoperation System}
\subsubsection{Robot System Setup}

The robot system consists of a bimanual wheeled robot equipped with two dexterous hands. The robot has 2 DoF in the legs, 3 DoF in the waist, 7 DoF in each arm, and 3 DoF in the head. It is equipped with two wrist cameras with an RGB resolution of 640$\times$528 and one head-mounted camera with an RGB resolution of 1280$\times$800. Each dexterous hand has 10 actuated DoF and 16 DoF in total. The thumb has 3 DoF; the index, ring, and little fingers each have 1 DoF for abduction/adduction; and the index, middle, ring, and little fingers each have 1 DoF for extension/flexion. 

\subsection{Teleoperation System Setup}

\begin{figure}[h]
    \centering
    \includegraphics[trim=0 0 0 0, clip, width=1.0\linewidth]{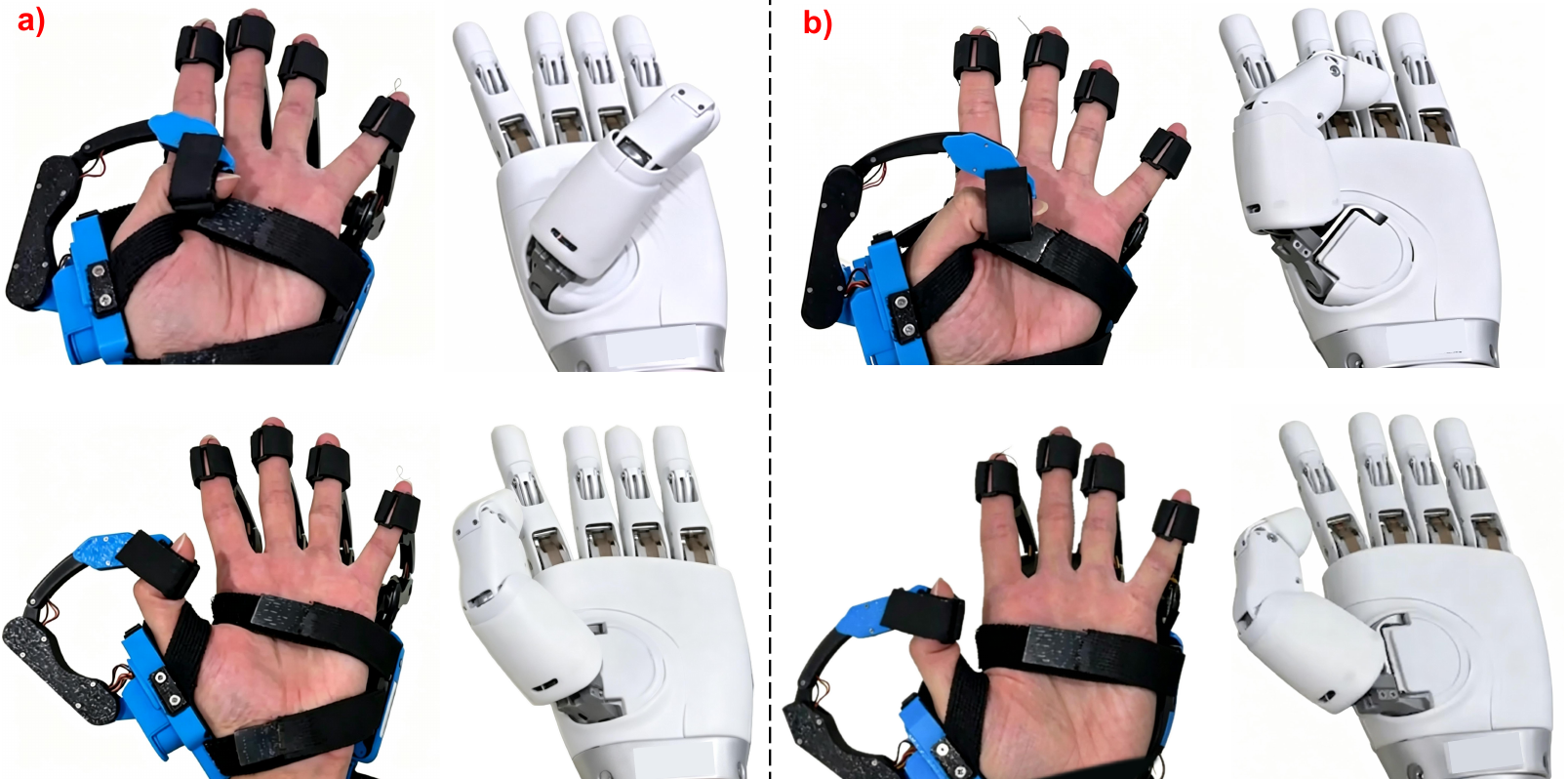}
    \caption{Visualization of different retargeting methods: a) DexPilot. b) Ours.
    }
    \label{fig:retarget}
\end{figure}

The teleoperation system consists of a pair of skeleton-based gloves for controlling the dexterous hands and a pair of trackers for controlling the robotic arms, with a paddle used to control the teleoperation signals. We read the joint angles of the gloves and apply forward kinematics to obtain the fingertip poses, which are then retargeted to the robot hand frame. The DexPilot~\cite{handa2020dexpilot} algorithm is used for inverse kinematics. However, the original algorithm forces the fingers to close when the distance between fingertips falls below a certain threshold, which affects fine-grained manipulation. Therefore, we disable this mechanism. On the other hand, the original algorithm only optimizes fingertip positions without considering orientations. As a result, when the human thumb is tilted, the robot thumb may remain upright, as shown in Fig.~\ref{fig:retarget}(a), which is less intuitive for teleoperation. To address this issue, we add an orientation constraint and jointly optimize fingertip positions and orientations, the optimized results are shown in Fig.~\ref{fig:retarget}(b).

\subsection{Training Recipe}

\subsubsection{Visual-Tactile Joint Pre-training}
Visual and tactile inputs are integrated within a unified pre-training framework. Initially, both modalities are standardized to a resolution of 224$\times$224 and processed using an identical tokenization strategy, which partitions the raw inputs into sequences of non-overlapping 196 tokens. To preserve spatial topology, fixed two-dimensional sine-cosine positional embeddings are added to the tokens of each modality. An independent random masking strategy is then applied to each modality with a uniform masking ratio of 75\%. The unmasked tokens are subsequently processed by modality-specific Vision Transformer (ViT) encoders. Following this encoding phase, we introduce learnable modality embeddings to explicitly distinguish the input sources and capture modality-specific characteristics. The extracted features from both modalities are then fed into a shared fusion encoder to facilitate multi-modal feature interaction. Finally, the fused representations are routed to their respective modality-specific decoders for output reconstruction.

\subsubsection{Sparse-Global Tactile Token Representation}
During policy training, each hand's tactile image is encoded by the same pre-trained tactile encoder into 196 spatial patch tokens plus one learnable tactile token prepended to the sequence. This learnable token aggregates information from every patch and produces a global summary of the full tactile field, serving a similar function to the classification (CLS) token~\cite{rao2021dynamicvit}, while each patch token retains localized contact. All tactile tokens are projected by an MLP to match the VLM backbone dimension. We then apply the contact gate only to the spatial patch tokens: active patch tokens and the per-hand global tactile token remain attendable, while inactive patch tokens are masked. These tactile tokens are concatenated with visual and language tokens to form the prefix tokens. Action tokens attend causally to these prefix tokens during the denoising process. In the VLM backbone, the learnable tactile token is prevented from attending to visual, language, and action tokens, while these tokens are allowed to attend to it. This asymmetric design preserves the aggregated global tactile feature while enabling other modalities to access tactile information when needed.

\subsubsection{Sparse Future Tactile Prediction}
We fix the number of predicted tactile frames to five for both SFTP and its DFTP control. SFTP uses stride $k{=}10$ over the action horizon $H{=}50$, yielding $T_{\mathrm{S}}=\{10,20,30,40,50\}$. DFTP instead predicts the five adjacent future frames $T_{\mathrm{D}}=\{1,2,3,4,5\}$. Thus, the two variants match the number of supervised targets while allocating them over different temporal extents. Since our tactile sensor only perceives normal force, and the normal force is assigned to the third channel when constructing the tactile image, we extract this channel and partition it into non-overlapping patches as the prediction target for each future tactile image of each hand. The prediction head projects each token from the prefix output, excluding the learnable tactile token, into the target force predictions.

\subsection{Training Details}
For tactile encoder pre-training, we adopt the MAE~\cite{he2022masked} architecture for visual-tactile joint pre-training, following M2VTP~\cite{liu2024masked}, and implement it in JAX/Flax. We select tactile images containing non-zero tactile signals, along with their corresponding RGB images from our self-collected 200-hour dataset, for pre-training. We set the batch size to 1024 and train for 15{,}000 iterations, corresponding to approximately one epoch. For policy training, we adopt the $\pi_{0.5}$~\cite{black2025pi_} architecture, implemented in JAX/Flax, consisting of a PaliGemma Gemma-2B vision-language backbone and a Gemma-300M action expert. The full input sequence comprises 1{,}562 tokens. Three camera images are encoded by a SigLIP So400m/14 image encoder into 256 tokens per view at 224$\times$224 resolution, contributing 768 tokens in total. The tactile images uses the pre-trained tactile encoder, producing 196 patch tokens plus 1 learnable tactile token per hand and 394 tokens for both hands, projected into the backbone dimension via a linear layer. Following $\pi_{0.5}$, the 64-dimensional proprioceptive state is discretized and serialized as text within the language prompt, tokenized by PaliGemma SentencePiece with a maximum sequence length of 350 tokens. The action suffix consists of 50 tokens corresponding to the action horizon $H{=}50$.

Training follows a two-stage pipeline. In the finetuning stage, we train for 6 epochs on 200 hours of multi-task dexterous manipulation data. The VLM backbone and action expert are initialized from a $\pi_{0.5}$ checkpoint. The original checkpoint targets a 32-dimensional action space, so we discard the action input/output projection layers and re-initialize them for our 64-dimensional space. Of the 64 dimensions, 34 are active: 7 each for the left and right end-effector poses, and 10 each for the left and right hand joints, with the remainder zero-padded. The tactile encoder starts from self-supervised pretrained weights and is finetuned with the full model. In the post-training stage, we train for 100 epochs on task-specific demonstrations, starting from the finetuned checkpoint.

\subsection{Sub-task Description}
We divide each task into several subtasks to evaluate the task progress of each method.

\textbf{Earbud flipping:} grasp the earbud, rotate it via in-hand manipulation, and release the earbud.

\textbf{Stacked-book retrieval:} pull out the target book, grasp the target book, safely retrieve it from the stack, and place it down.

\textbf{Postcard retrieval:} slide the postcard to the edge of the table, grasp the postcard, and place it into the box.

\textbf{Multi-object grasping:} each hand sequentially grasp two objects while maintaining the grasp of the previously acquired object, and then dispose of the two objects into the trash can one by one. Successful grasping and successful disposal are counted as separate subtasks for each object, resulting in eight subtasks in total.

Although earbud flipping, stacked-book retrieval, and postcard retrieval are sequential tasks, the later subtasks in multi-object grasping do not necessarily depend on the successful completion of earlier subtasks. Therefore, the task completion rate (TCR) is calculated as the percentage of completed subtasks. For example, in the multi-object grasping task, the robot may fail to grasp the first object and consequently fail to dispose of it, while successfully completing all remaining subtasks. In this case, the subtask completion sequence would be 01011111, resulting in a task completion rate of 0.75 for that trial.

\subsection{Implementation Details of Baselines}
We unify the input and output representations across all baseline models to be consistent with \textit{Ours}. Specifically, for \textit{LDA-1B}, we additionally incorporate two wrist-view images and modify its state and action representations accordingly. For \textit{GR00T N1.7}, we adopt the same three-view visual input, consisting of two wrist views and one top view, and modify its state and action representations accordingly. Both baseline models use an action prediction horizon of $H{=}50$, the representation of end-effector poses and hand joint positions is consistent with \textit{Ours}, with rotation represented as quaternions. Apart from these modifications, all other hyperparameters follow the default settings of their respective models. All models are initialized from the officially released pretrained weights and use a batch size of 256 during post-training. For the additional finetuning experiments, we use a batch size of 512.

\subsection*{Acknowledgments}
We thank Ken Chen, Qijun Chen, Ziyu Zhang, Zhihao Chen, Linfeng Zhao, and Yongyi Wan for their assistance in setting up the real-world robotic platform and teleoperation system. We also thank Ningyuan Huang for his support in setting up the baseline environments.

\end{document}